\documentclass[letterpaper, 10 pt, conference]{ieeeconf}
\IEEEoverridecommandlockouts   % This command is only needed if 
\usepackage{graphics} % for pdf, bitmapped graphics files
\usepackage{epsfig} % for postscript graphics files
\usepackage{caption}

\usepackage{amsmath}
\usepackage{amssymb}
\usepackage{pifont}

\DeclareMathOperator*{\argmax}{arg\,max}
\usepackage{mathabx}  % for \widecheck

\usepackage{booktabs}
\usepackage{multirow}
\usepackage[dvipsnames,table,xcdraw]{xcolor}
\usepackage{soul} % for \hl

\makeatletter
\let\NAT@parse\undefined
\makeatother
\usepackage{hyperref}
\hypersetup{
    colorlinks,breaklinks,
    linkcolor=[RGB]{255 0 0},
    citecolor=[RGB]{31 81 255},
    urlcolor=[RGB]{0 0 0}
}

\title{\LARGE \bf 
GeoGS-SLAM: Online Monocular Reconstruction Using Gaussian Splatting with Geometric Priors
}

\author{Ruilan Gao$^{1}$, Letian Jin$^{1}$, Yu Zhang$^{1,2,*}$%
\thanks{This work was supported by the National Natural Science Foundation of China (Grant No. 62576311), in part by NSFC 62088101 Autonomous Intelligent Unmanned Systems, and in part by Zhejiang Provincial Natural Science Foundation of China under Grant No. LD24F030001.}%
\thanks{$^{1}$State Key Laboratory of Industrial Control Technology, College of Control Science and Engineering, Zhejiang University, Hangzhou, China, 310027.}%
\thanks{$^{2}$Key Laboratory of Collaborative Sensing and Autonomous Unmanned Systems of Zhejiang Province, Hangzhou, China, 310027.}%
\thanks{$^{*}$Corresponding author: Yu Zhang (Email: \href{mailto:zhangyu80@zju.edu.cn}{zhangyu80@zju.edu.cn}).}%
}

\newcommand{\insertfig}{
    \centering
    \vspace{6pt}
    \includegraphics[width=0.93\linewidth]{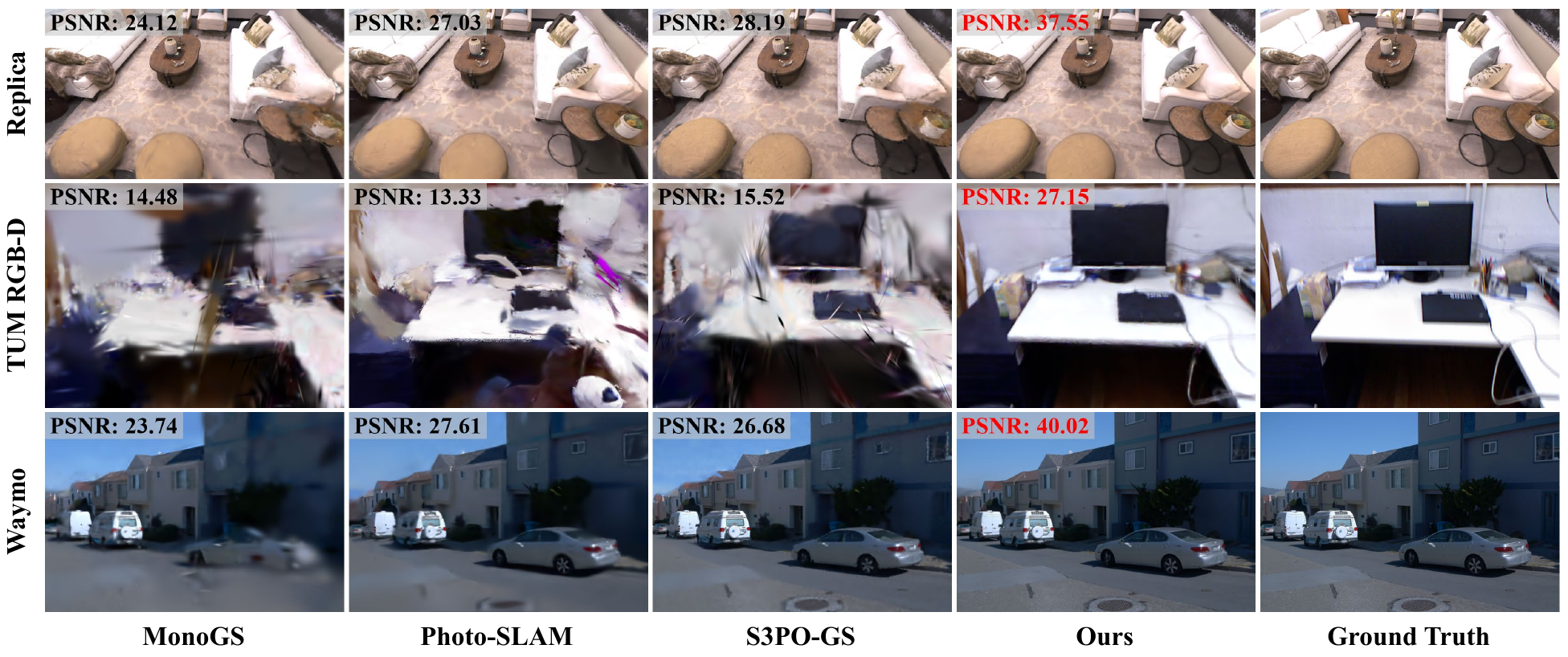}
    \setcounter{figure}{0}
    \captionof{figure}{%
    \textbf{Rendering results on three datasets.}
    Our method produces high-fidelity reconstructions on both indoor and outdoor benchmarks, outperforming state-of-the-art monocular 3DGS-based SLAM methods.
    }
    \label{fig:vis_beginning}
    \vspace{-3pt}
}

\begin{document}

\makeatletter
\apptocmd{\@maketitle}{\insertfig}{}{}% insert the figure after authors
\makeatother

\maketitle

\thispagestyle{empty}
\pagestyle{empty}

%%%%%%%%%%%%%%%%%%%%%%%%%%%%%%%%%%%%%%%%%%%%
\begin{abstract}
%%% 3DGS SLAM
SLAM methods based on 3D Gaussian Splatting (3DGS) have demonstrated impressive tracking and mapping performance, but typically require additional geometric information from external depth sensors.
%%% FF prior SLAM
Meanwhile, recent SLAM systems that leverage geometric priors from pre-trained feed-forward models enable real-time dense reconstruction, yet often discard original RGB information during optimization, thus degrading overall reconstruction quality.
%%% ours
We present GeoGS-SLAM, an online monocular dense reconstruction system that combines the 3DGS-based map representation with learned geometric priors.
Given uncalibrated RGB input, we first employ a feed-forward visual geometry model to predict camera and scene priors.
The Gaussian scene map is then expanded by directly sampling Gaussian primitives from both RGB input and geometric priors.
Camera poses and the scene map are jointly optimized through a coarse-to-fine strategy that minimizes both photometric and geometric losses.
To ensure global consistency, we further incorporate online loop closure detection and pose graph optimization.
%%% results
Extensive experiments across indoor and outdoor benchmarks demonstrate that GeoGS-SLAM achieves superior rendering quality and tracking accuracy compared to state-of-the-art methods while maintaining online real-time performance.
% 
%%%%%%%%%%%%%%%%%%%%%%%%%%%%%%%%%%%%%%%%%%
Project page: \url{https://rlgao.github.io/geogs_slam}.
%%%%%%%%%%%%%%%%%%%%%%%%%%%%%%%%%%%%%%%%%%
\end{abstract}

%%%%%%%%%%%%%%%%%%%%%%%%%%%%%%%%%%%%%%%%%%%%
\section{INTRODUCTION}

%%% SLAM
Simultaneous localization and mapping (SLAM) is a core problem in computer vision, serving as the foundation for applications ranging from robotics and autonomous driving to augmented reality systems.
%%% 2 trends
Recent advances in SLAM have been propelled by two complementary breakthroughs in view synthesis and 3D reconstruction: radiance field rendering \cite{tosi2024NerfGsSlamSurvey} and feed-forward scene reconstruction \cite{zhang2025FF3DReconstructionSurvey}.

%%% 3DGS SLAM
The advent of neural radiance field (NeRF) \cite{mildenhall2021nerf} and 3D Gaussian Splatting (3DGS) \cite{kerbl20233DGS} has fundamentally transformed scene representations in SLAM systems. 
In particular, 3DGS employs differentiable rasterization of 3D Gaussians to achieve efficient, photorealistic rendering, and has been shown to support high-quality tracking and mapping in SLAM \cite{matsuki2024MonoGS, keetha2024splatam}.
However, existing 3DGS-based SLAM systems predominantly rely on external geometric measurements from depth sensors and exhibit degraded performance when constrained to RGB-only input \cite{yu2025OpenGSSLAM}.

%%% FF prior SLAM
More recently, feed-forward models such as DUSt3R \cite{wang2024dust3r} and VGGT \cite{wang2025vggt} have revolutionized 3D scene reconstruction through Transformer-based architectures trained at scale. 
A growing number of SLAM systems now leverage geometric priors from these powerful models to achieve real-time pose estimation and dense scene reconstruction \cite{murai2025MASt3R-SLAM, liu2025slam3r}. 
However, these methods typically treat the learned priors as the primary optimization signal while excluding original RGB observations from the optimization loop.
For example, VGGT-SLAM \cite{maggio2025Vggt-slam} relies on the alignment of point map priors predicted by VGGT. 
This limitation of discarding photometric information prevents closed-loop verification against visual evidence, potentially degrading reconstruction fidelity and consistency.

%%% ours
In this paper we present GeoGS-SLAM, an online monocular dense reconstruction system that synergistically combines 3DGS-based map representation with learned geometric priors. 
The key idea is to construct a closed-loop reconstruction pipeline that leverages feed-forward priors for geometric bootstrapping while preserving original RGB evidence for radiance field rendering-based optimization.

Our approach begins by employing a pre-trained visual geometry model to predict camera intrinsics and extrinsics, depth maps, and point maps from uncalibrated RGB input. 
Followed by scale alignment, these priors provide robust multi-view geometric guidance. 
We then directly sample Gaussian primitives from both images and geometric priors to expand the 3D Gaussian scene map. 
The map and camera poses are jointly refined through a coarse-to-fine, rendering-based optimization that minimizes both photometric and geometric losses. 
To ensure global consistency, we further integrate online loop closure detection and pose graph optimization.

%%% results
Comprehensive experiments across indoor and outdoor benchmarks, 
including Replica \cite{straub2019replica}, TUM RGB-D \cite{sturm2012TUMRGB-D} and Waymo \cite{sun2020Waymo}, 
demonstrate that GeoGS-SLAM achieves superior rendering quality and tracking accuracy 
compared to state-of-the-art (SOTA) monocular SLAM methods 
while maintaining online real-time performance.

%%% contributions
The main contributions of our work are:
\begin{itemize}
    \item We propose GeoGS-SLAM, a novel RGB-only SLAM system that integrates 3D Gaussian Splatting with feed-forward geometric priors in a unified closed-loop reconstruction pipeline, achieving robust tracking and high-fidelity mapping.
    \item We leverage efficient modules, including direct primitive sampling, rendering-based joint optimization, and online loop closure to enable accurate pose estimation and photorealistic reconstruction with real-time processing capabilities.
    \item We provide extensive experimental validation across indoor and outdoor benchmarks, demonstrating the superior performance of our approach compared to SOTA monocular SLAM methods in tracking accuracy and rendering quality.
\end{itemize}

%%%%%%%%%%%%%%%%%%%%%%%%%%%%%%%%%%%%%%%%%%%%
\section{RELATED WORK}

%%%%%%%%%%%%%%%%%%%%%%%%
\subsection{Classical Visual SLAM}

%%% feature-based
Early visual SLAM methods mainly employ feature-based pipelines.  % \cite{davison2007monoslam}.
PTAM \cite{klein2007PTAM} introduces the first parallelized tracking and mapping framework, using sparse feature correspondences and bundle adjustment (BA) to produce accurate trajectories and sparse 3D maps.
ORB-SLAM \cite{mur2015ORB-SLAM} and its successors \cite{mur2017Orb-slam2, campos2021Orb-slam3} extend this paradigm with efficient feature extraction, loop closure detection, and pose graph optimization to reduce drift and maintain long-term consistency.

%%% direct
In contrast to feature-based approaches, direct methods such as LSD-SLAM \cite{engel2014LSD-SLAM} and DSO \cite{engel2017DSO} operate directly on pixel intensities rather than extracted keypoints, offering enhanced robustness in texture-poor scenes at the cost of increased photometric sensitivity. 
Dense SLAM methods, often incorporating multi-sensor configurations (stereo or RGB-D), enable richer scene reconstruction, producing dense maps suitable for interaction and navigation \cite{newcombe2011kinectfusion,dai2017bundlefusion}.
% \cite{newcombe2011kinectfusion,dai2017bundlefusion,newcombe2011dtam}

%%% limitations
Nonetheless, sparse feature-based methods provide robust tracking but only sparse geometry, while dense methods yield richer maps yet are photometrically sensitive and computationally expensive. 
These limitations have motivated recent efforts to integrate photorealistic rendering or learned priors into SLAM frameworks.

%%%%%%%%%%%%%%%%%%%%%%%%
\subsection{Radiance Field-based SLAM}

%%% NeRF
Scene representations in SLAM systems have undergone a paradigm shift with the emergence of NeRF \cite{mildenhall2021nerf,Orbeez-SLAM,HERO-SLAM}.
iMAP \cite{sucar2021imap} has pioneered the integration of neural implicit representations into SLAM, utilizing multilayer perceptrons (MLPs) to encode both geometry and appearance within a unified framework.
NICE-SLAM \cite{zhu2022nice-slam} and Vox-Fusion \cite{yang2022vox-fusion} further extend this paradigm by incorporating hierarchical feature grids and voxel-based neural implicit surface representations to enhance reconstruction quality and computational efficiency.

%%% 3DGS
More recently, 3DGS \cite{kerbl20233DGS} has emerged as a compelling alternative to NeRF-based representations, offering real-time rendering through differentiable rasterization of 3D Gaussians \cite{Hier-SLAM,OpenGS-SLAM-Semantic,yan2024Gs-slam}.
Pioneering methods such as MonoGS \cite{matsuki2024MonoGS} and SplaTAM \cite{keetha2024splatam} demonstrate the successful integration of 3DGS as the sole scene representation for SLAM, achieving robust frame-to-model tracking and mapping through joint optimization of camera poses and Gaussian primitives.
Other approaches, including Photo-SLAM \cite{huang2024photo-slam}, incorporate separate tracking modules to improve pose estimation accuracy.
Recent works have focused on extending 3DGS-based SLAM to large-scale outdoor environments \cite{yu2025OpenGSSLAM, xin2025large-ScaleGSSLAM, cheng2025S3PO-GS}.

%%% limitations
However, most existing radiance field-based SLAM systems rely heavily on depth measurements from RGB-D sensors, limiting their applicability in scenarios where only monocular RGB input is available.
Therefore, we leverage pre-trained geometric priors to enable robust bootstrapping and optimization of 3DGS-based SLAM systems operating on uncalibrated RGB-only input.

%%%%%%%%%%%%%%%%%%%%%%%%
\subsection{Geometric Prior-based SLAM}

The recent advancement of feed-forward reconstruction models \cite{wang2024dust3r,wang2025vggt,leroy2024mast3r,yang2025fast3r} has introduced a novel paradigm in SLAM through learned geometric priors, enabling dense reconstruction without relying on traditional geometric pipelines.
MASt3R-SLAM \cite{murai2025MASt3R-SLAM} leverages MASt3R-predicted point maps and matching features \cite{leroy2024mast3r} to construct a real-time dense monocular SLAM system, achieving globally consistent pose estimation and dense reconstruction.
Similarly, SLAM3R \cite{liu2025slam3r} is built upon DUSt3R \cite{wang2024dust3r}, utilizing the Image-to-Points (I2P) and Local-to-World (L2W) modules to establish an end-to-end dense reconstruction framework.
In contrast to these two-view prior-based approaches, VGGT-SLAM \cite{maggio2025Vggt-slam} leverages the more powerful VGGT architecture \cite{wang2025vggt}, employing point map alignment strategies for dense RGB SLAM.

However, these geometric prior-based SLAM methods predominantly use the learned priors as the primary optimization signal, where original RGB observations are often discarded.
This limitation prevents closed-loop verification against photometric evidence and can compromise reconstruction quality.
Therefore, we integrate geometric priors with photorealistic rendering-based optimization to achieve both accurate tracking and high-fidelity reconstruction.

%%%%%%%%%%%%%%%%%%%%%%%%%%%%%%%%%%%%%%%%%%%%
\section{METHOD}

\begin{figure*}[htbp]
    \centering
    \includegraphics[width=0.86\linewidth]{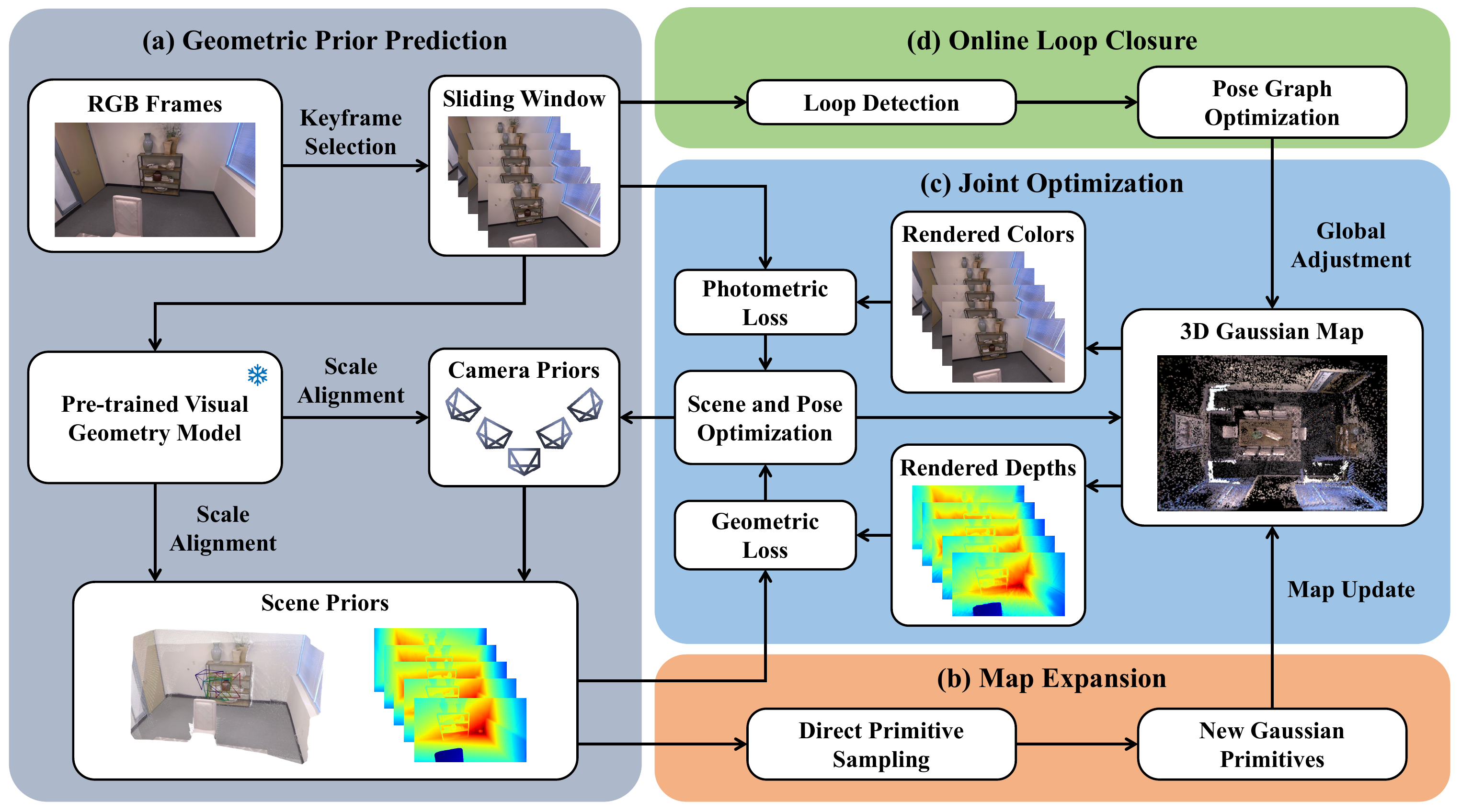}
%%%%%%%%%%%%%%%%%%%%%%%%%%%%%%%%%%%%
    % \vspace{-5pt}
    \caption{%
    \textbf{Overview of GeoGS-SLAM.} 
    (a) Given uncalibrated RGB input, the system first selects keyframes and performs \textit{Geometric Prior Prediction}, producing camera and scene priors using a pre-trained visual geometry model. 
    (b) \textit{Map Expansion} directly samples Gaussian primitives from both input images and geometric priors to update the map. 
    (c) The map and camera poses are refined through rendering-based \textit{Joint Optimization} that minimizes both photometric and geometric losses. 
    (d) \textit{Online Loop Closure} followed by pose graph optimization is integrated to further enhance global consistency.
    }
%%%%%%%%%%%%%%%%%%%%%%%%%%%%%%%%%%%%
    \label{fig:overview}
\vspace{-10pt}
\end{figure*}

GeoGS-SLAM integrates 3D Gaussian Splatting with learned geometric priors in a unified framework for monocular dense reconstruction, as illustrated in Fig. \ref{fig:overview}. 

% ==================================================
% ==================================================

Our approach consists of four core components: 
\textit{Geometric Prior Prediction} (Sec. \ref{subsec:GeometricPriorPrediction}) leverages a pre-trained model to produce priors of camera parameters and scene geometry from RGB keyframes; 
\textit{Map Expansion} (Sec. \ref{subsec:MapExpansion}) updates the Gaussian map through direct primitive sampling; 
\textit{Joint Optimization} (Sec. \ref{subsec:JointOptimization}) refines both the map and poses via rendering-based photometric and geometric loss minimization; 
and \textit{Online Loop Closure} (Sec. \ref{subsec:OnlineLoopClosure}) with pose graph optimization further ensures global consistency.

%%%%%%%%%%%%%%%%%%%%%%%%
\subsection{3DGS Scene Representation}

%%% Gaussian param
We employ 3DGS as the scene representation, where each anisotropic Gaussian primitive $\mathcal{G}_i$ ($i=1,\ldots,N$) is parameterized by the following properties: 
%%%
position $\boldsymbol{\mu}_i \in \mathbb{R}^3$, 
color $\mathbf{c}_i \in \mathbb{R}^3$, 
scale $\mathbf{S}_i =\text{diag}(\mathbf{s}_i) \in \mathbb{R}^{3 \times 3}$, 
rotation $\mathbf{R}_i \in \mathbb{R}^{3 \times 3}$, 
and 
opacity $o_i \in [0,1]$.
%%%
The covariance matrix $\boldsymbol{\Sigma}_i$ defining the ellipsoidal shape is computed as 
\begin{equation}
    \boldsymbol{\Sigma}_i = \mathbf{R}_i \mathbf{S}_i \mathbf{S}_i^\top \mathbf{R}_i^\top.
\end{equation}

%%% render - splat
To render color and depth images from a given world-to-camera pose 
$\mathbf{T} = 
\begin{bmatrix}
    \mathbf{R} & \mathbf{t} \\
    \mathbf{0}         & 1
\end{bmatrix}
\in \mathrm{SE}(3)$, 
the 3D Gaussians are first projected onto the image plane, with each splatted 2D Gaussian $\mathcal{G}_i'(\boldsymbol{\mu}_i',\boldsymbol{\Sigma}_i')$ obtained via 
\begin{equation}
    \boldsymbol{\mu}_i' = \pi(\mathbf{R} \boldsymbol{\mu}_i + \mathbf{t}),
    \,
    \boldsymbol{\Sigma}_i' = \mathbf{J} \mathbf{R} \boldsymbol{\Sigma}_i \mathbf{R}^\top \mathbf{J}^\top,
\end{equation}
where $\pi(\cdot)$ denotes the projection operation, 
and $\mathbf{J}$ is the Jacobian of the affine approximation of the projective transformation.

%%% render - color, depth
The final color and depth at pixel $\mathbf{x}'$ can be rendered through $\alpha$-blending of the depth-sorted Gaussians, given by 
\begin{equation}
    \mathbf{C} = \sum_{i=1}^N \mathbf{c}_i \alpha_i \prod_{j=1}^{i-1} (1-\alpha_j),
    \,
    D = \sum_{i=1}^N d_i \alpha_i \prod_{j=1}^{i-1} (1-\alpha_j),
\end{equation}
where the opacity weight $\alpha_i$ for each Gaussian is computed via 
\begin{equation}
    \alpha_i = o_i \exp \left( 
        -\frac{1}{2} 
        (\mathbf{x}'-\boldsymbol{\mu}_i')^\top 
        \left(\boldsymbol{\Sigma}_i'\right)^{-1}
        (\mathbf{x}'-\boldsymbol{\mu}_i')
    \right).
\end{equation}

%%%%%%%%%%%%%%%%%%%%%%%%
\subsection{Geometric Prior Prediction}
\label{subsec:GeometricPriorPrediction}

%%% why VGGT
We leverage the visual geometry model VGGT \cite{wang2025vggt} for geometric prior prediction, 
which processes image sets of arbitrary length and predicts comprehensive 3D attributes in a single forward pass. 
This choice addresses the limitations of earlier two-view models like DUSt3R \cite{wang2024dust3r} and MASt3R \cite{leroy2024mast3r}, 
which are constrained to pairwise inference and thus limit multi-view consistency.

%%% sliding window formation
Similar to VGGT-SLAM \cite{maggio2025Vggt-slam}, we employ a sliding window strategy to organize keyframes for multi-view prediction. 
An uncalibrated RGB frame is selected as a keyframe and added to the active sliding window 
if its optical flow-based relative displacement from the last keyframe exceeds a predefined threshold $\tau_\text{disp}$. 
The initial window $\mathcal{W}^1$ accumulates keyframes $\mathbf{I}_1^1, \ldots, \mathbf{I}_w^1$ until reaching the fixed window size $w$. 
For subsequent windows $\mathcal{W}^k$ ($k>1$), we use the last keyframe from the previous window as the first keyframe,  
i.e., $\mathbf{I}_w^{k-1} = \mathbf{I}_0^k$, 
and add new keyframes $\mathbf{I}_1^k, \ldots, \mathbf{I}_w^k$ until it contains $w+1$ frames in total. 
This overlapping design facilitates scale alignment between consecutive windows, as discussed below.

%%% predict
Each window $\mathcal{W}^k$ containing keyframes $\mathbf{I}_i^k \in \mathbb{R}^{W \times H \times 3}$ ($i=0,\ldots,w$) 
is processed by the pre-trained model to generate predictions $\mathcal{F}(\mathcal{W}^k)$, 
including camera extrinsics $\widehat{\mathbf{T}}_i^k \in \mathrm{SE}(3)$, 
intrinsics $\widehat{\mathbf{K}}_i^k \in \mathbb{R}^{3 \times 3}$, 
depth maps $\widehat{\mathbf{D}}_i^k \in \mathbb{R}^{W \times H}$, 
and corresponding confidence maps $\widehat{\mathbf{Q}}_i^k \in \mathbb{R}^{W \times H}$. 
To obtain point maps $\widehat{\mathbf{X}}_i^k \in \mathbb{R}^{W \times H \times 3}$ expressed in each keyframe's coordinate system, 
we unproject the depth maps to 3D using the predicted camera parameters rather than employing direct point map regression through the DPT head, 
as this approach yields better accuracy according to the original findings \cite{wang2025vggt}.

%%% scale alignment
We align the scale of current predictions $\mathcal{F}(\mathcal{W}^k)$ 
with previous ones $\mathcal{F}(\mathcal{W}^{k-1})$ 
using the mutual keyframe $\mathbf{I}_w^{k-1} = \mathbf{I}_0^k$. 
For each pixel $(u,v)$ where both predictions exhibit high confidence 
(i.e., $\widehat{\mathbf{Q}}_w^{k-1}(u,v) > \tau_{\text{conf}}$ 
and $\widehat{\mathbf{Q}}_0^{k}(u,v) > \tau_{\text{conf}}$), 
we compute the scale factor $\rho_{k-1,k}$ by 
\begin{equation}
    \rho_{k-1,k} = \frac{1}{\vert \mathcal{V} \vert} 
        \sum_{(u,v) \in \mathcal{V}}
        \frac{
            \left\Vert \widehat{\mathbf{X}}_w^{k-1}(u,v) \right\Vert 
        }{
            \left\Vert \widehat{\mathbf{X}}_0^{k}(u,v) \right\Vert 
        },
\end{equation}
where $\mathcal{V}$ denotes the set of valid high-confidence pixels. 
This scale factor is applied to adjust all metric predictions in $\mathcal{F}(\mathcal{W}^k)$, 
including translational components $\widehat{\mathbf{t}}_i^k$ of camera poses $\widehat{\mathbf{T}}_i^k$, 
depth maps $\widehat{\mathbf{D}}_i^k$, 
and point maps $\widehat{\mathbf{X}}_i^k$ ($i=0,\ldots,w$), 
thereby ensuring scale consistency across all geometric priors.

%%%%%%%%%%%%%%%%%%%%%%%%
\subsection{Map Expansion}
\label{subsec:MapExpansion}

%%% P1
We leverage a direct sampling strategy to generate new Gaussian primitives from both input images and predicted geometric priors, 
inspired by recent work \cite{meuleman2025onthefly}.
This approach enables map expansion while avoiding redundant primitive placement through a two-stage probability assessment.
To determine optimal locations for new Gaussian primitives, 
we utilize the Difference of Gaussians (DoG) operator \cite{marr1980theory-edge-detection} to identify regions with rich geometric detail.
The probability matrix $\mathbf{P}_1 \in \mathbb{R}^{W \times H}$ for primitive placement at each pixel in image $\mathbf{I}$ is computed as 
%%%
% https://en.wikipedia.org/wiki/Difference_of_Gaussians#cite_note-2
%%%
\begin{equation}
\label{eq:DoG}
    \mathbf{P}_1 = \left\Vert
        \left( 
            \mathbf{\Phi}_{\sigma_1} - \mathbf{\Phi}_{\sigma_2}
        \right) * \mathbf{I}
    \right\Vert,
\end{equation}
where $\mathbf{\Phi}_{\sigma}$ denotes a Gaussian kernel with zero mean and standard deviation $\sigma$, 
and we set $\sigma_1 = 0.5, \sigma_2 = 1.5$.

%%% P2
To prevent redundant primitive placement in already well-represented regions, 
we render a synthetic view $\widetilde{\mathbf{I}}$ using the current 3D Gaussian map from the predicted camera pose, 
and compute a corresponding occupancy probability $\mathbf{P}_2$ using the same DoG operator as in Eq. (\ref{eq:DoG}). 
A new Gaussian primitive can be spawned at pixel $(u,v)$ only when 
the difference between placement and occupancy probabilities exceeds threshold $\tau_{\text{prim}}$, 
i.e., $\mathbf{P}_1(u,v) - \mathbf{P}_2(u,v) > \tau_{\text{prim}}$.

%%% primitive properties
For each qualified pixel $(u,v)$, 
we initialize the new Gaussian primitive by combining information from both the input image and geometric priors. 
The color $\mathbf{c}$ is directly sampled from the corresponding pixel value $\mathbf{I}(u,v)$. 
The 3D position $\boldsymbol{\mu}$ is obtained from the point prior $\widehat{\mathbf{X}}(u,v)$ transformed to the world coordinate system.
The scale $s$ for each dimension is computed based on 
the predicted focal length $\widehat{f}$ from intrinsics $\widehat{\mathbf{K}}$ 
and the point position $\widehat{\mathbf{X}}(u,v)$ in the camera coordinate system,
as given by 
\begin{equation}
    s = \frac{\left\Vert \widehat{\mathbf{X}}(u,v) \right\Vert}{\widehat{f}} + \epsilon,
\end{equation}
where $\epsilon$ is a constant.
% , following established practices \cite{wu2025monocular}.

%%%%%%%%%%%%%%%%%%%%%%%%
\subsection{Joint Optimization}
\label{subsec:JointOptimization}

Following Gaussian primitive expansion from the current keyframe, 
we perform online joint optimization of both the scene map $\mathcal{G}$ and camera poses within the active keyframe window $\mathcal{W}$ 
by minimizing a combination of photometric and geometric losses. 
% 
%%% photometric loss
The photometric loss $\mathcal{L}_{\text{pho}}$ enforces visual consistency 
between rendered and observed images by combining L1 and SSIM losses \cite{wang2004SSIM},
as computed by
\begin{equation}
    \mathcal{L}_{\text{pho}} 
    = \left\Vert \widetilde{\mathbf{I}} - \mathbf{I} \right\Vert_1
        + \lambda_{\text{SSIM}} \left(
            1 - \text{SSIM}( \widetilde{\mathbf{I}}, \mathbf{I} )
        \right),
\end{equation}
where $\widetilde{\mathbf{I}}$ represents the rendered image 
and $\mathbf{I}$ denotes the input ground truth.

%%% geometric loss
The geometric loss $\mathcal{L}_{\text{geo}}$ maintains consistency between 
the rendered depth map $\widetilde{\mathbf{D}}$ and the predicted depth prior $\widehat{\mathbf{D}}$,
as obtained by
\begin{equation}
    \mathcal{L}_{\text{geo}} = \left\Vert \widetilde{\mathbf{D}} - \widehat{\mathbf{D}} \right\Vert_1.
\end{equation}

%%% joint optimization
The complete optimization objective combines photometric and geometric terms across all keyframes in the active window,
which is formulated as 
\begin{equation}
    \min_{\mathcal{G}, \{\mathbf{T}_i\}} 
    \sum_{\mathbf{I}_i \in \mathcal{W}} 
    \left( 
        \mathcal{L}_{\text{pho}, i} + \lambda_{\text{geo}} \mathcal{L}_{\text{geo}, i} 
    \right).
\end{equation}

%%% photometric loss
To accelerate convergence and improve optimization stability, 
we employ a coarse-to-fine training strategy \cite{huang2024photo-slam}. 
We construct $n$-level image pyramids for both the input image and prior depth map of each keyframe, 
where level $l$ corresponds to the image or depth map downsampled by a factor of $2^l$.
Optimization begins at the coarsest level using downsampled images and depth maps as the photometric and geometric supervision signals, respectively, 
then progressively decreasing the pyramid level at fixed iteration intervals until reaching the original full resolution. 
This multi-scale training approach enables rapid initial convergence while preserving fine-grained details in the final optimization stages.

%%%%%%%%%%%%%%%%%%%%%%%%
\subsection{Online Loop Closure}
\label{subsec:OnlineLoopClosure}

To mitigate accumulated drift and ensure global consistency, we incorporate online loop closure detection with pose graph optimization. 

%%% loop closure detection
We first extract a global descriptor $\mathbf{d} \in \mathbb{R}^{8448}$ for each keyframe 
using the pre-trained MegaLoc model \cite{berton2025Megaloc}, which provides robust place recognition capability.
For a query keyframe $\mathbf{I}_{\text{query}}^k$, 
loop closure detection proceeds by identifying the most similar keyframe candidate $\mathbf{I}_{\text{cand}}^l$ from all previous windows 
based on the cosine similarity between the global descriptors, as obtained via  
\begin{equation}
    \mathbf{I}_{\text{cand}}^l
    = \argmax_{\mathbf{I}_i^j \in \mathcal{W}^j,\, \forall j<k}
    \cos \left( \mathbf{d}(\mathbf{I}_{\text{query}}^k), \mathbf{d}(\mathbf{I}_i^j) \right).
\end{equation}

To prevent false positive detections, the candidate is validated as a genuine loop frame $\mathbf{I}_{\text{loop}}^l$ 
only if the similarity score exceeds a predefined threshold $\tau_{\text{sim}}$, as given by 
\begin{equation}
    \mathbf{I}_{\text{loop}}^l = 
    \begin{cases}
        \mathbf{I}_{\text{cand}}^l, 
            & \text{if } 
            \cos \left( \mathbf{d}(\mathbf{I}_{\text{query}}^k), \mathbf{d}(\mathbf{I}_{\text{cand}}^l) \right) 
            > \tau_{\text{sim}}, \\
        \text{null}, & \text{otherwise}.
    \end{cases}
\end{equation}

%%% pose graph optimization
Upon detecting valid loop closures, we extend the active window to include all detected loop frames, as obtained via 
\begin{equation}
    \overline{\mathcal{W}}^k = 
    \{ \mathbf{I}_0^k, \ldots, \mathbf{I}_w^k \}
    \cup
    \{ \mathbf{I}_{\text{loop},1}^{l_1}, \ldots, \mathbf{I}_{\text{loop},m}^{l_m} \}.
\end{equation}

This extended window is processed by the feed-forward geometry model to obtain relative pose estimates 
between query frames in $\overline{\mathcal{W}}^k$ and 
corresponding loop frames from previous windows $\mathcal{W}^{l_1}, \ldots, \mathcal{W}^{l_m}$.
After scale alignment, these pose estimates serve as loop closure constraints in the pose graph, 
complementing the odometry constraints derived from locally optimized keyframe poses. 
Then the complete pose graph is optimized using the Levenberg-Marquardt algorithm, 
yielding pose correction 
$
\Delta\mathbf{T}_i = 
\begin{bmatrix}
    \Delta\mathbf{R}_i & \Delta\mathbf{t}_i \\
    \mathbf{0}         & 1
\end{bmatrix}
$ 
for each keyframe $\mathbf{I}_i$.

%%% pose, map update
Finally, we apply these pose corrections to update both keyframe poses and 
the associated Gaussian primitives, as given by 
\begin{gather}
    \mathbf{T}_i'                     = \Delta\mathbf{T}_i \mathbf{T}_i, \\
    \boldsymbol{\mu}_{\mathcal{G}_i}' = \Delta\mathbf{R}_i \boldsymbol{\mu}_{\mathcal{G}_i} + \Delta\mathbf{t}_i, 
    \,
    \mathbf{R}_{\mathcal{G}_i}'       = \Delta\mathbf{R}_i \mathbf{R}_{\mathcal{G}_i}, 
\end{gather}
where $\mathcal{G}_i$ is a primitive originally spawned by keyframe $\mathbf{I}_i$.

%%%%%%%%%%%%%%%%%%%%%%%%%%%%%%%%%%%%%%%%%%%%
\section{EXPERIMENTS}

%%%%%%%%%%%%%%%%%%%%%%%%
\subsection{Experimental Setup}

%%%%%%
i) \textbf{Datasets.}
We conduct experiments across indoor and outdoor benchmarks, including Replica \cite{straub2019replica}, TUM RGB-D \cite{sturm2012TUMRGB-D} and Waymo \cite{sun2020Waymo}.
Specifically, we use the \texttt{Office} and \texttt{Room} sequences from Replica, 
the \texttt{fr1} sequences from TUM RGB-D, 
and nine 200-frame sequences from Waymo.
Results presented below are averaged across sequences within each respective dataset.

%%%%%%
ii) \textbf{Metrics.}
We use PSNR, SSIM \cite{wang2004SSIM} and LPIPS \cite{zhang2018LPIPS} for rendering quality evaluation, 
and the RMSE of ATE for tracking accuracy assessment.

%%%%%%
iii) \textbf{Baseline Methods.}
For rendering quality evaluation, our method is compared against SOTA radiance field-based SLAM methods operating on monocular RGB input, 
including neural implicit representation-based approaches 
(%
GO-SLAM \cite{zhang2023go-slam},
GlORIE-SLAM \cite{zhang2024glorie-slam}%
) 
and 3DGS-based methods 
(%
MonoGS \cite{matsuki2024MonoGS},
Photo-SLAM \cite{huang2024photo-slam},
Splat-SLAM \cite{sandstrom2025splat-slam},
DROID-Splat \cite{homeyer2024DROID-Splat},
S3PO-GS \cite{cheng2025S3PO-GS}%
).
%%%
For tracking accuracy assessment, we additionally compare against geometric prior-based SLAM methods 
(%
SLAM3R \cite{liu2025slam3r},
MASt3R-SLAM \cite{murai2025MASt3R-SLAM},
VGGT-SLAM \cite{maggio2025Vggt-slam}%
).

%%%%%%
iv) \textbf{Implementation Details.}
Our method is implemented using PyTorch and CUDA.
All experiments are conducted on a desktop with an NVIDIA RTX 4090 GPU and an Intel Core i9-13900K CPU.

%%%%%%%%%%%%%%%%%%%%%%%%
\subsection{Results}

%%%%%%
i) \textbf{Rendering Quality.}
%%% table 
Tab. \ref{tab:all-render} presents rendering quality comparisons across all evaluated datasets, showing that our method achieves SOTA performance across all metrics and benchmarks.
Compared to current best radiance field-based monocular SLAM methods, our method delivers substantial improvements in PSNR: \textbf{+4.94} on Replica, \textbf{+3.15} on TUM RGB-D, and \textbf{+3.81} on Waymo.

\begin{table*}[ht]
\centering
\setlength{\tabcolsep}{8pt}
\renewcommand{\arraystretch}{1}

\caption{%
\textbf{Rendering quality comparison across three datasets.}
Best results are highlighted in \textbf{bold}. 
Our method demonstrates superior rendering performance compared to both 
neural implicit representation-based approaches (\textit{NI}) 
and 3DGS-based methods (\textit{GS}) across all indoor and outdoor benchmarks.
}
% \vspace{-5pt}
\label{tab:all-render}

\begin{tabular}{@{}clccccccccc@{}}
\toprule
\multicolumn{1}{l}{}                              & \multirow{2}{*}{Method}                   & \multicolumn{3}{c}{Replica}                                              & \multicolumn{3}{c}{TUM RGB-D}                                            & \multicolumn{3}{c}{Waymo}                           \\ \cmidrule(l){3-11} 
\multicolumn{1}{l}{}                              &                                           & PSNR$\uparrow$ & SSIM$\uparrow$ & \multicolumn{1}{c|}{LPIPS$\downarrow$} & PSNR$\uparrow$ & SSIM$\uparrow$ & \multicolumn{1}{c|}{LPIPS$\downarrow$} & PSNR$\uparrow$ & SSIM$\uparrow$ & LPIPS$\downarrow$ \\ \midrule
\multicolumn{1}{c|}{\multirow{2}{*}{\textit{NI}}} & GO-SLAM \cite{zhang2023go-slam}           & 24.20          & 0.935          & \multicolumn{1}{c|}{0.462}             & 12.59          & 0.601          & \multicolumn{1}{c|}{0.727}             & 9.72           & 0.109          & 0.812             \\
\multicolumn{1}{c|}{}                             & GlORIE-SLAM \cite{zhang2024glorie-slam}   & 24.73          & 0.911          & \multicolumn{1}{c|}{0.171}             & 17.20          & 0.725          & \multicolumn{1}{c|}{0.384}             & 23.90          & 0.892          & 0.247             \\ \midrule
\multicolumn{1}{c|}{\multirow{6}{*}{\textit{GS}}} & MonoGS \cite{matsuki2024MonoGS}           & 26.77          & 0.848          & \multicolumn{1}{c|}{0.281}             & 15.27          & 0.534          & \multicolumn{1}{c|}{0.575}             & 16.60          & 0.649          & 0.655             \\
\multicolumn{1}{c|}{}                             & Photo-SLAM \cite{huang2024photo-slam}     & 32.62          & 0.917          & \multicolumn{1}{c|}{0.162}             & 16.46          & 0.593          & \multicolumn{1}{c|}{0.471}             & 19.86          & 0.767          & 0.595             \\
\multicolumn{1}{c|}{}                             & Splat-SLAM \cite{sandstrom2025splat-slam} & 31.86          & 0.906          & \multicolumn{1}{c|}{0.133}             & 20.87          & 0.700          & \multicolumn{1}{c|}{0.370}             & 25.10          & 0.773          & 0.349             \\
\multicolumn{1}{c|}{}                             & DROID-Splat \cite{homeyer2024DROID-Splat} & 29.96          & 0.882          & \multicolumn{1}{c|}{0.227}             & 19.80          & 0.685          & \multicolumn{1}{c|}{0.397}             & 22.84          & 0.737          & 0.350             \\
\multicolumn{1}{c|}{}                             & S3PO-GS \cite{cheng2025S3PO-GS}           & 29.94          & 0.892          & \multicolumn{1}{c|}{0.199}             & 18.00          & 0.627          & \multicolumn{1}{c|}{0.442}             & 21.82          & 0.798          & 0.471             \\
\multicolumn{1}{c|}{}                             & \textbf{GeoGS-SLAM (Ours)}                & \textbf{37.56} & \textbf{0.968} & \multicolumn{1}{c|}{\textbf{0.039}}    & \textbf{24.02} & \textbf{0.797} & \multicolumn{1}{c|}{\textbf{0.220}}    & \textbf{28.91} & \textbf{0.900} & \textbf{0.135}    \\ \bottomrule
\end{tabular}

% \vspace{-2pt}
\end{table*}

\begin{figure*}[htbp]
    \centering
    \includegraphics[width=0.93\linewidth]{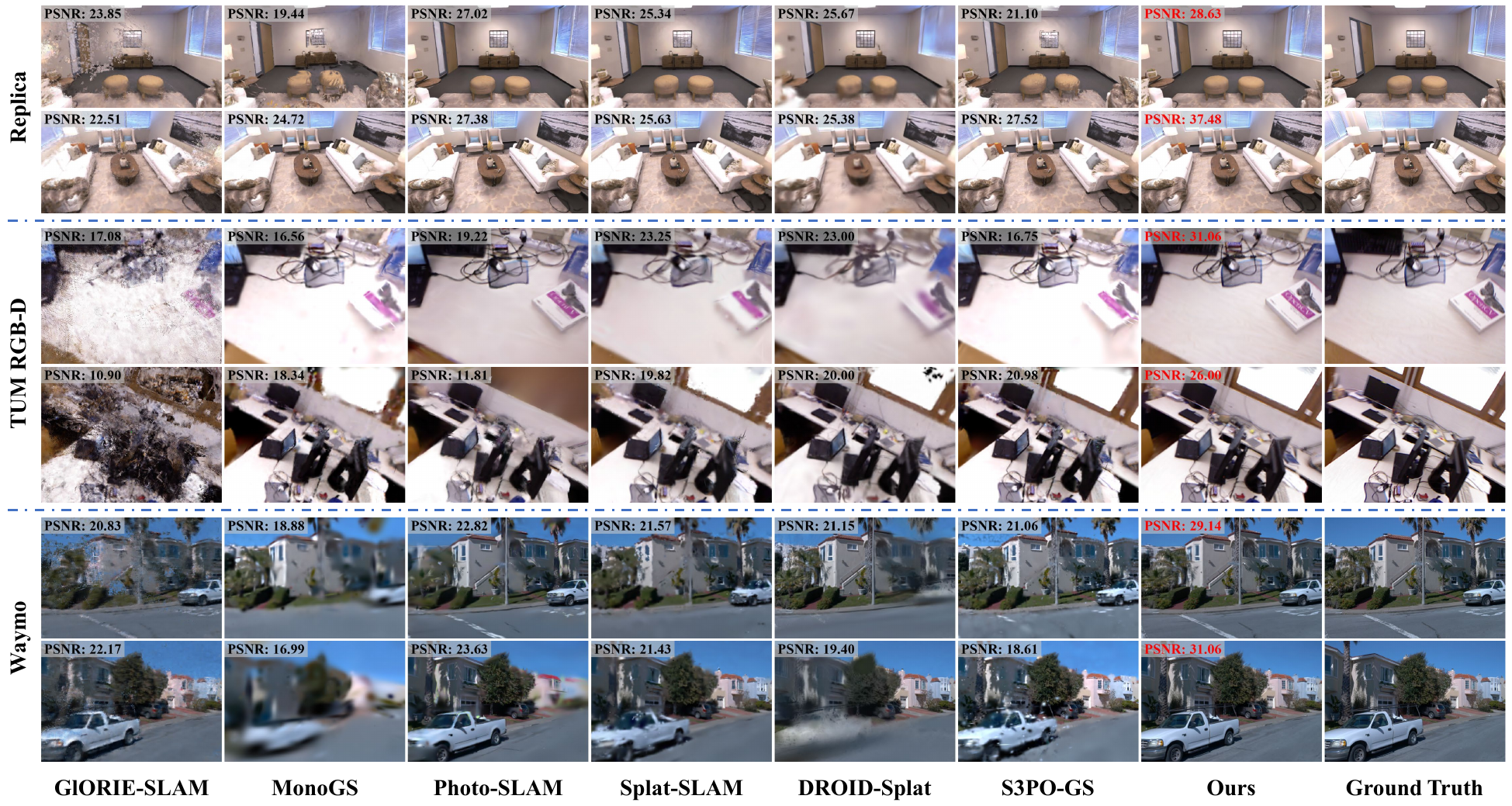}
%%%%%%%%%%%%%%%%%%%%%%%%%%%%%%%%%%%%%%%%%%%%%%%%
    % \vspace{-3pt}
    \caption{%
    \textbf{Comparison of rendering results.}
    Our method produces photorealistic reconstructions for both indoor and outdoor scenes. 
    For indoor environments, it captures fine-grained textures and geometric details with minimal artifacts.
    For outdoor scenarios, it successfully handles complex driving scenes, preserving architectural structures and vehicle details.
    }
%%%%%%%%%%%%%%%%%%%%%%%%%%%%%%%%%%%%%%%%%%%%%%%%
    \label{fig:vis_results}
    \vspace{-10pt}
\end{figure*}

%%% figure
Fig. \ref{fig:vis_results} provides qualitative comparisons across representative scenes from all three datasets. 
For indoor environments (Replica and TUM RGB-D), our method produces detailed reconstructions that accurately capture fine-grained textures, lighting variations, and geometric structures. 
For complex outdoor scenarios (Waymo), our method successfully handles challenging conditions including varying illumination, dynamic content, and large-scale environments, showcasing its scalability to real-world autonomous driving applications.

These results demonstrate the substantial benefits of integrating learned geometric priors into 3DGS-based SLAM frameworks.

\begin{figure*}[htbp]
    \centering
    \includegraphics[width=0.93\linewidth]{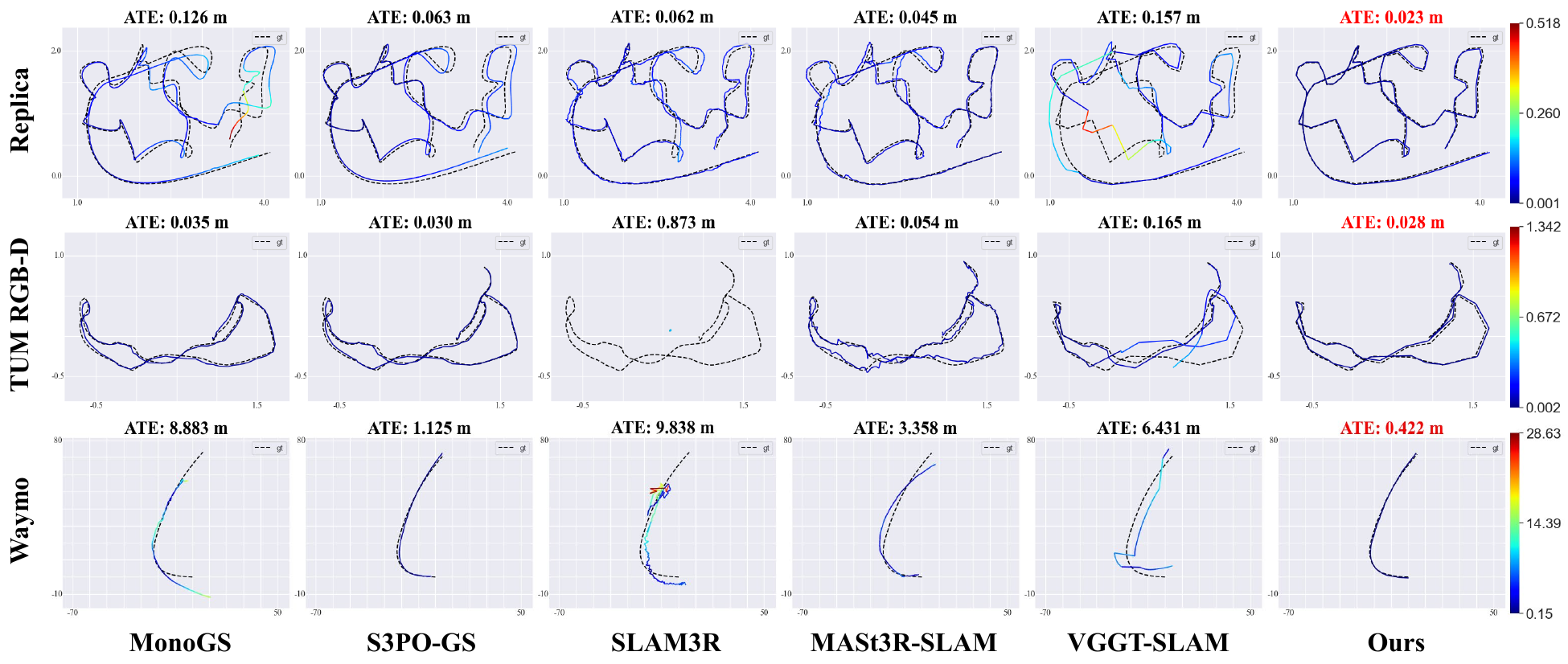}
%%%%%%%%%%%%%%%%%%%%%%%%%%%%%%%%%%%%%%%%%%%%%%%%
    % \vspace{-3pt}
    \caption{%
    \textbf{Comparison of estimated trajectories.}
    Each tracking result is projected onto the $x$-$y$ plane, with ground truth shown as a dashed line.
    Our method achieves superior accuracy and robustness across indoor and outdoor environments.
    }
%%%%%%%%%%%%%%%%%%%%%%%%%%%%%%%%%%%%%%%%%%%%%%%%
    \label{fig:vis_ate}
    \vspace{-5pt}
\end{figure*}

%%%%%%
ii) \textbf{Tracking Accuracy.}
%%% table 
Tab. \ref{tab:all-track} presents tracking results across the datasets,
with the comparison of estimated trajectories shown in Fig. \ref{fig:vis_ate}.
Our method achieves SOTA tracking performance compared to methods operating on uncalibrated RGB input, while maintaining competitive accuracy against approaches that leverage calibrated camera parameters.

\begin{table}[ht]
\centering
\setlength{\tabcolsep}{5pt}
\renewcommand{\arraystretch}{1}

\caption{%
\textbf{Tracking accuracy comparison across three datasets.}
ATE RMSE [m] ($\downarrow$) is reported.
% , with best results highlighted in \textbf{bold}. 
Our method achieves superior tracking performance compared to methods operating on uncalibrated RGB input (\textit{Uncalib.}), 
while maintaining comparable accuracy against approaches utilizing calibrated camera parameters (\textit{Calib.}).
}
% \vspace{-5pt}
\label{tab:all-track}

% 9B9B9B

\begin{tabular}{@{}clccc@{}}
\toprule
                                                                     & Method                                                           & Replica                               & TUM RGB-D                             & Waymo                                 \\ \midrule
\multicolumn{1}{c|}{{\color[HTML]{9B9B9B} }}                         & {\color[HTML]{9B9B9B} GlORIE-SLAM \cite{zhang2024glorie-slam}}   & {\color[HTML]{9B9B9B} 0.039}          & {\color[HTML]{9B9B9B} \textbf{0.046}} & {\color[HTML]{9B9B9B} 0.500}          \\
\multicolumn{1}{c|}{{\color[HTML]{9B9B9B} }}                         & {\color[HTML]{9B9B9B} MonoGS \cite{matsuki2024MonoGS}}           & {\color[HTML]{9B9B9B} 0.348}          & {\color[HTML]{9B9B9B} 0.302}          & {\color[HTML]{9B9B9B} 7.077}          \\
\multicolumn{1}{c|}{{\color[HTML]{9B9B9B} }}                         & {\color[HTML]{9B9B9B} Photo-SLAM \cite{huang2024photo-slam}}     & {\color[HTML]{9B9B9B} 0.022}          & {\color[HTML]{9B9B9B} 0.325}          & {\color[HTML]{9B9B9B} \textbf{0.366}} \\
\multicolumn{1}{c|}{{\color[HTML]{9B9B9B} }}                         & {\color[HTML]{9B9B9B} Splat-SLAM \cite{sandstrom2025splat-slam}} & {\color[HTML]{9B9B9B} \textbf{0.018}} & {\color[HTML]{9B9B9B} 0.052}          & {\color[HTML]{9B9B9B} 0.495}          \\
\multicolumn{1}{c|}{\multirow{-5}{*}{{\color[HTML]{9B9B9B} \textit{Calib.}}}} & {\color[HTML]{9B9B9B} S3PO-GS \cite{cheng2025S3PO-GS}}           & {\color[HTML]{9B9B9B} 0.075}          & {\color[HTML]{9B9B9B} 0.121}          & {\color[HTML]{9B9B9B} 0.869}          \\ \midrule
\multicolumn{1}{c|}{}                                                & SLAM3R \cite{liu2025slam3r}                                      & 0.064                                 & 0.652                                 & 18.265                                \\
\multicolumn{1}{c|}{}                                                & MASt3R-SLAM \cite{murai2025MASt3R-SLAM}                          & 0.045                                 & 0.059                                 & 2.431                                 \\
\multicolumn{1}{c|}{}                                                & VGGT-SLAM \cite{maggio2025Vggt-slam}                             & 0.177                                 & 0.121                                 & 7.517                                 \\
\multicolumn{1}{c|}{\multirow{-4}{*}{\textit{Uncalib.}}}             & \textbf{GeoGS-SLAM (Ours)}                                       & \textbf{0.024}                        & \textbf{0.035}                        & \textbf{0.861}                        \\ \bottomrule
\end{tabular}

\vspace{-3pt}
\end{table}

%%% figure

Specifically, our method outperforms the best-performing geometric prior-based SLAM method, MASt3R-SLAM, reducing tracking error by \textbf{46.7\%} on Replica, \textbf{40.7\%} on TUM RGB-D, and \textbf{64.6\%} on Waymo. 
While these methods, like MASt3R-SLAM and VGGT-SLAM, achieve reasonable initial pose estimates through learned priors, they fail to refine these estimates against actual image evidence.
This limitation leads to accumulated drift and reduced accuracy over extended sequences and large-scale environments with greater scene complexity.

These substantial tracking improvements across diverse environments validate the effectiveness of our closed-loop optimization approach that fuses geometric prior guidance with photometric refinement.

%%%%%%
iii) \textbf{Real-time Performance.}
%%% table 
We conduct runtime analysis as presented in Tab. \ref{tab:runtime}.
The results demonstrate that our system achieves real-time performance by applying joint optimization only at keyframes.
Our selective optimization strategy enables efficient online reconstruction while maintaining high-quality tracking and mapping performance.

\begin{table}[ht]
\centering
\setlength{\tabcolsep}{12pt}
\renewcommand{\arraystretch}{1}

\caption{%
\textbf{Runtime evaluation results.}
Average processing time per keyframe across datasets is reported, demonstrating the computational efficiency of our approach.
}
% \vspace{-5pt}
\label{tab:runtime}

\begin{tabular}{@{}lc@{}}
\toprule
Step (per keyframe)                 & Time {[}ms{]}  \\ \midrule
Keyframe decision                   & 5.8            \\
Prior prediction                    & 57.2           \\
Scale alignment                     & 0.5            \\
Primitive sampling \& Map expansion & 9.2            \\
Joint optimization                  & 480.6          \\
Loop closure detection              & 6.5            \\
Pose graph optimization             & 0.6            \\
Global map adjustment               & 21.7           \\ \midrule
\textbf{Sum}                        & \textbf{582.1} \\ \bottomrule
\end{tabular}

\vspace{-2pt}
\end{table}

%%%%%%%%%%%%%%%%%%%%%%%%
\subsection{Ablation Study}

To validate the contribution of each component in our system, we conduct comprehensive ablation studies as presented in Tab. \ref{tab:ablation_replica}.

\begin{table}[ht]
\centering
\setlength{\tabcolsep}{9pt}
\renewcommand{\arraystretch}{1}

\caption{%
\textbf{Ablation study results on Replica.}
% Best results are highlighted in \textbf{bold}. 
Each row shows the impact of removing a component from the system.
}
% \vspace{-5pt}
\label{tab:ablation_replica}

\begin{tabular}{@{}lcccc@{}}
\toprule
Method                 & PSNR$\uparrow$ & SSIM$\uparrow$ & LPIPS$\downarrow$ & ATE$\downarrow$ \\ \midrule
w/o loop closure       & 37.18          & 0.967          & 0.041             & 0.034           \\
% w/o pose optimization  & 30.49          & 0.912          & 0.079             & 0.029           \\
w/o primitive sampling & 35.62          & 0.954          & 0.065             & \textbf{0.024}  \\
w/o scene priors    & 22.28          & 0.799          & 0.345             & 0.052           \\
w/o camera priors       & 15.94          & 0.673          & 0.671             & 0.922           \\ \midrule
\textbf{Ours}           & \textbf{37.56} & \textbf{0.968} & \textbf{0.039}    & \textbf{0.024}  \\ \bottomrule
\end{tabular}

\vspace{-5pt}
\end{table}

%%%
i) \textbf{Loop Closure.}
Removing loop closure detection and pose graph optimization 
% (w/o loop closure) 
leads to significant degradation in tracking accuracy, 
demonstrating that our system's pose estimation substantially benefits from global consistency enhancement.

%%%
ii) \textbf{Primitive Sampling.}
Replacing our direct primitive sampling strategy with uniform sampling approaches employed in most previous works \cite{yan2024Gs-slam} 
% (w/o primitive sampling) 
results in notable rendering quality deterioration, confirming the effectiveness of our sampling method.

%%%
iii) \textbf{Scene Priors.}
Eliminating feed-forward scene priors and using randomly initialized depth values as adopted by previous monocular methods \cite{matsuki2024MonoGS} 
% (w/o scene priors) 
causes severe decline in both rendering and tracking performance, validating the importance of geometric constraints for monocular 3DGS-based SLAM.

%%%
iv) \textbf{Camera Priors.}
Most critically, removing learned camera priors 
and instead using raw intrinsic estimates with the last keyframe's pose for new frame initialization 
% (w/o camera priors) 
leads to catastrophic degradation in both rendering and tracking performance. 
This result underscores the essential role of learned camera priors in addressing the inherent scale ambiguity and convergence challenges of monocular SLAM.

%%%%%%%%%%%%%%%%%%%%%%%%%%%%%%%%%%%%%%%%%%%%
\section{CONCLUSIONS}

In this paper we propose GeoGS-SLAM, a novel monocular SLAM system that integrates 3DGS with learned geometric priors. 
Our method achieves SOTA performance against existing monocular SLAM systems based on radiance fields and geometric priors across various benchmarks. 
Our work establishes a new paradigm for visual SLAM by building a closed-loop pipeline that leverages feed-forward priors for geometric bootstrapping while preserving photometric evidence for radiance field rendering-based optimization to achieve high-fidelity online reconstruction.

%%%%%%%%%%%%%%%%%%%%%%%%%%%%%%%%%%%%%%%%%%%%
\bibliographystyle{IEEEtran} 
\bibliography{ref}

\end{document}